\documentclass[conference,compsoc]{IEEEtran}
\usepackage{times}
\usepackage{helvet}
\usepackage{courier}
\usepackage{url}
\usepackage{graphicx}
\usepackage{amsmath}
\usepackage{amssymb}
\DeclareMathOperator*{\argmax}{arg~max}

\ifCLASSOPTIONcompsoc
  \usepackage[nocompress]{cite}
\else
  \usepackage{cite}
\fi
\ifCLASSINFOpdf
\else
\fi
\begin{document}
%
\title{Detecting Table Region in PDF Documents Using Distant Supervision}

\author{\IEEEauthorblockN{Miao Fan}
\IEEEauthorblockA{
Tsinghua University\\
Beijing, 100084, China\\
fanmiao.cslt.thu@gmail.com}
\and
\IEEEauthorblockN{Doo Soon Kim}
\IEEEauthorblockA{
Bosch Research\\
Palo Alto, CA, 95014, U.S.A.\\
doosoon.kim@us.bosch.com }
}


%


\maketitle

\begin{abstract}
Superior to state-of-the-art approaches which compete in table recognition with 67 annotated government reports in PDF format released by {\it ICDAR 2013 Table Competition}, this paper contributes a novel paradigm leveraging large-scale unlabeled PDF documents to open-domain table detection. We integrate the paradigm into our latest developed system ({\it PdfExtra}) to detect the region of tables by means of 9,466 academic articles from the entire repository of {\it ACL Anthology}, where almost all papers are archived by PDF format without annotation for tables. The paradigm first designs heuristics to automatically construct weakly labeled data. It then feeds diverse evidences, such as layouts of documents and linguistic features, which are extracted by {\it Apache PDFBox} and processed by {\it Stanford NLP} toolkit, into different canonical classifiers. We finally use these classifiers, i.e. {\it Naive Bayes}, {\it Logistic Regression} and {\it Support Vector Machine}, to collaboratively vote on the region of tables. Experimental results show that {\it PdfExtra} achieves a great leap forward, compared with the state-of-the-art approach. Moreover, we discuss the factors of different features, learning models and even domains of documents that may impact the performance. Extensive evaluations demonstrate that our paradigm is compatible enough to leverage various features and learning models for open-domain table region detection within PDF files.
\end{abstract}


%
\IEEEpeerreviewmaketitle

\section{Introduction}

Tables are primarily used to present data such as the results of statistical analysis, experimental records, attributes of items, etc. The grid structure of the table -- columns and rows -- allows a reader to easily interpret and compare different items. Due to the advantages, tables have been widely adopted in many different articles such as web pages, academic publications, online manuals. Computer scientists who conduct research on information extraction take delight in engaging with tables that occur in those electronic documents, as they are the natural sources to feed and populate relational databases.

Some formats of the electronic documents are machine-readable, such as {\it HTML}, {\it XML} and even {\it TEX}. These formats derive from {\it SGML} (Standard Generalized Markup Language) and inherit the basic principle that the language pins a pair of specific tags to mark a snippet of text.
For example, {\it HTML} files use $\langle table \rangle$ as the start and $\langle /table \rangle$ as the end, to indicate the region of a table. AI programs can easily recognize expected regions with the help of tags, and extract the information that we want with pre-defined actions. 
However, it is tedious for our human beings to read the markup language, because we are sensitive to the layouts of documents, and focus more on the contents. Therefore, the {\it Portable Document Format} (PDF) was designed as a file format to represent a document independent of the platform it displays, and to preserve the layouts both on screen and in print. These strengths draw much attention from the online publishing. So far, many academic papers and manuals have adopted PDF as the standard format.

Unfortunately, we meet Waterloo when detecting the region of tables within PDF files, due to the lack of structural information. To the best of our knowledge, the latest off-the-shelf software, {\it Apache PDFBox}\footnote{\url{https://pdfbox.apache.org/}}, could only provide the coordinates $(x, y)$ and the font style of each character in a PDF document. As table region detection is the fundamental and significant step for further information extraction from PDF files, fruitful approaches have been proposed in recent decades. However, they either simply design heuristic rules based on pre-defined layouts, or adopt supervised learning techniques fed by few annotated corpora from restricted domains. For instance, {\it ICDAR 2013} set up a competition about table detection and structure recognition within 67 annotated PDF documents posted by U.S. and E.U. governments, where each document is accompanied by a {\it XML} file to indicate the location of tables.

When we further apply these methods to some free access digital academic archives, such as {\it IEEE Xplore} and {\it Springer Link}, the variety of layouts and explosive amount of unannotated data expose the urgent demand on unsupervised or semi-supervised frameworks. By means of these frameworks, we do not have to spend much labor on annotation, but can leverage large-scale unlabeled PDF files. To the best of our knowledge, Klampfl et al. \cite{klampfl2014comparison} have recently proposed unsupervised table recognition methods applied on digital scientific articles. However, their work was purely based on heuristic rules and evaluated on 109 files in total. We consider it not flexible enough to handle more PDF articles with variable layouts.

Therefore, we firstly propose a novel solution which requires very little human efforts in detecting a table region in PDF. Specifically, our approach reduces the cost of training data annotation by automatically generating the annotated data using a {\it distant supervision} technique\cite{fan2014distant,fan2015distant}. The approach first collects a large amount of unlabeled PDF dataset, uses simple heuristic rules to automatically annotate the unlabeled dataset and then train a supervised classifer over the (weakly-labeled) training examples to predict the boundary of table region. The human efforts are almost neglegible in our approach because the unlabeled PDF dataset can be easily acquired from the web and the data annotation is automatically performed.

Our experimental results confirm the promise of our approach. To evaluate our approach, we collected 9,446 PDF files from {\it ACL Anthology}\footnote{\url{http://aclweb.org/anthology/}} and developed a simple heuristic rule to automatically generate training examples from them. Then, a supervised classifier (an ensenble of {\it Logistic Regression}, {\it Support Vector Machine} and {\it Naive Bayes}) was trained over the weakly-labeled datasets to be applied over the test datasets from several different domains. Our evaluation shows that, first, our approach significantly outperforms a state-of-the art algorithm \cite{klampfl2014comparison} for the ACL test dataset. Furthermore, even for a out-of-domain test dataset (ICDAR 2013 competition dataset \cite{gobel2013icdar}), our system achieve a significantly higher accuracy than the baseline system, indicating the effectiveness of a large amount of training dataset automatically generated in our approach. We also performed additional experiments to analyze the important features in detecting a table region and to compare different classifiers, and report the results of those experiments.


\section{Related Work}
A comprehensive review can be found in the final report of {\it ICDAR 2013 Table Competition} \cite{gobel2013icdar}, which announced the performances of recent academic and commercial systems on either table region detection or table structure analysis. Here we restrict our survey on a number of recent methods that attempt to discover table regions within PDF files.

The first effort was the {\it pdf2table}\footnote{\url{http://ieg.ifs.tuwien.ac.at/projects/pdf2table/}} system  \cite{Yildiz2005}, which used heuristics to detect the table region. It assumed that a table had more than one column, and a table region was formed by merging neighboring multi-lines. However, the algorithm could only handle pages with single-column layouts.

The {\it PDF-TREX} system \cite{Oro2009} considered a set of words as seeds first, and identified tables in a bottom-up manner. Specifically, words were aligned and grouped to lines based on their vertical overlap, and line segments were
obtained by applying hierarchical agglomerative clustering to the words. According to the number of segments, a line was categorized into three classes: text lines, table lines, and unknown lines. Then, the table region could be found by combining
contiguous table lines or unknown lines.

Supervised classification models were mainly adopted by Liu et al. \cite{Liu:2008:ITB:1458082.1458255}, who designed a table detection method that leveraged heuristics to construct lines from individual characters and to select those sparse lines that occur within a table for training. Starting from a table caption, these sparse lines were then iteratively merged to a table region. This approach is very similar to the state-of-the-art unsupervised method \cite{klampfl2014comparison} and ours, except that it was built upon labeled text blocks instead of lines.

The up-to-date approach \cite{klampfl2014comparison} did not rely on annotated data, but used complex heuristics to achieve comparable performances with supervise-based systems. Our system {\it PdfExtra} costs free on labeled data but covers large-scale PDF files with varies layouts. Therefore, we mainly compare the performance of system with the state-of-the-art unsupervised method \cite{klampfl2014comparison}.

\section{Paradigm}
{\it PdfExtra} benefits a lot from the off-the-shelf software {\it Apache PDFBox} which can recognize almost all characters within a PDF document. Beyond the characters, the software also provides the horizontal and vertical coordinates, as well as the font style for each of them. Thus each ``rich character'' can be represented as a tuple: $\langle character, x-axis, y-axis, font-type, font-size\rangle$. In addition, {\it Apache PDFBox} can merge the characters together into words, and return words in sequence that visually lay in the same line. There is nothing more that it can do to discover tables. Therefore, we leverage the outputs from {\it Apache PDFBox} and engage in predicting whether each line belongs to a table or not.

Although we have formulated the table region detection task into a binary classification problem, we still suffer the lack of annotated training data. As illustrated by Figure 1, the paradigm that we have designed to fix the issue contains three phases:
\begin{figure*}
\centering
\includegraphics[width=0.9\textwidth]{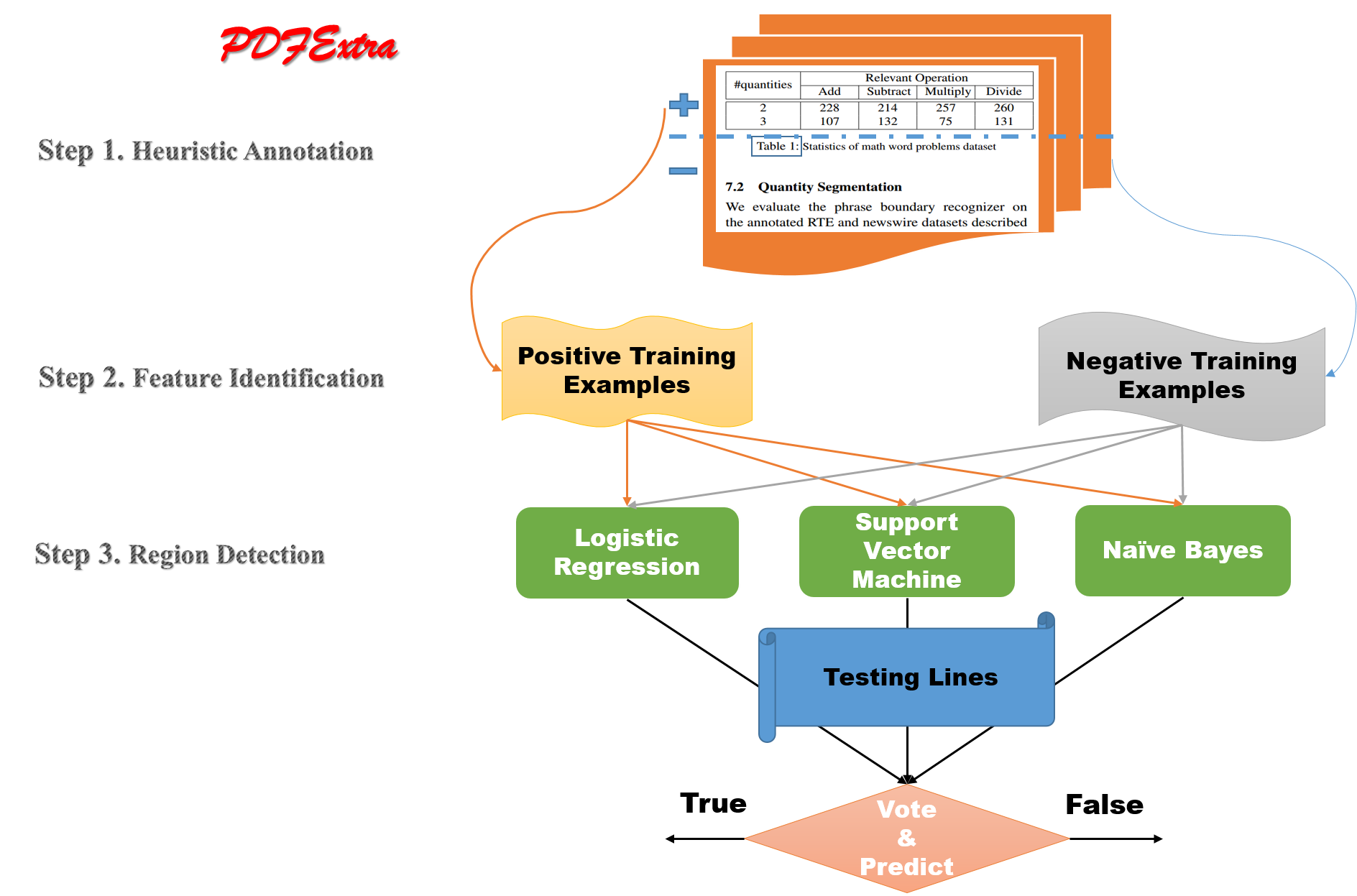}\\
\caption{The proposed paradigm adopted by the {\it PDFExtra} system.}

\end{figure*}
\subsection{Heuristic annotation}
Inspired by the idea of distant supervision \cite{fan2014distant,fan2015distant}, we adopt heuristics that can help automatically generate large-scale weakly labeled training examples.
More specifically,
we create a spider that downloads academic articles from {\it ACL Anthology}\footnote{\url{http://aclweb.org/anthology/}}, in which almost all papers are archived in PDF format. 9,466 literatures in total the year 2000 to 2015 are collected.
For a PDF article, we process each page in three steps as follows,
\begin{itemize}
\item {\it Indicator Recognition}: As all camera-ready drafts must conform to a limited number of official templates to be published, the word "Table" or "Tab." that appears in front of a line generally indicates the caption of a table. In other words, we find the lower or the upper boundary of the table region which depends on the templates.
\item {\it Surrounding Contexts}: The caption line plays a role in separating the table from the main body. Because we do not know which portion belongs to a table, we usually extend $k$ lines up and down as the candidate context.
\item {\it Positive v.s. Negative Examples}: After extracting these candidates, we assume that the group of lines with more blanks/margins will more likely locate in a table, rather than the other group. In this way, we can construct a balanced corpus for binary classification, even if it is weakly annotated by heuristics above.
\end{itemize}
By means of the heuristics we have proposed, a large-scale weakly labeled dataset can be automatically constructed for training. For instance, the rules help us prepare more than 350,000 lines as training examples extracted from {\it ACL Anthology}. As each line is a sequence of words in which every "rich character" with its coordinates and font style, we can further process each word to mark its start and end coordinates in the horizontal direction.

\subsection{Feature identification}
The state-of-the-art approach \cite{klampfl2014comparison} only concerns about the layouts of a PDF document. It iteratively includes a sparse block into a table in the buttom-up manner, where a block is identified as ``sparse'' if 1) their width is smaller than $\frac{2}{3}$ of the average width of a text block, or 2) there exists a gap between two consecutive words in the block that is larger than than two times the average width between two words in the document.

However, we believe that both linguistic and layout features are significant. Therefore, we select three kinds of features based on our observation that may contribute to detecting the region of tables. They are:
\begin{itemize}
\item {\it Normalized Average Margin} ({\it NAM}): According to the horizontal coordinate of each word in lines, we calculate the average margin between two consecutive words, so that each line is assigned by the feature. In most cases, the average margin between two consecutive words in the main body equals to the size of a space, and that in the tables usually occupies more. However, the average margin differs along with layouts, and generally the one-column layouts generate much larger margin than the two-column formats. Therefore, we normalize the average margin within the same page to be the feature that represents the layouts.
\item {\it POS Tag Distribution} ({\it PTD}): It is a common consensus that we prefer displaying information in a more structural and condensed way in tables, rather than flowery language expressed by sentences in the main body of an article. Intuitively, more noun phrases appear in tables, but less adjectives and adverbs are used. This distinction leads to the diverse distribution of the part-of-speech (POS) tags, which we concern as the second feature. There are 5 kinds of part-of-speech tags under our consideration processed by {\it Stanford POS Tagger\footnote{\url{http://nlp.stanford.edu/software/tagger.shtml}}}: {\it NN}, {\it VB}, {\it JJ}, {\it RB} and {\it OTHERS}.
\item {\it Named Entity Percentage} ({\it NEP}): We extend the traditional scope of named entities and include the number and the time. Therefore, 5 kinds of named entity tags, i.e. {\it PERSON}, {\it LOCATION}, {\it ORGANIZATION}, {\it NUMBER} and {\it TIME}, are recognized by {\it Stanford Named Entity Recognizer\footnote{\url{http://nlp.stanford.edu/software/CRF-NER.shtml}}}. For each kind of named entity, we compute its percentage in each line.
\end{itemize}
\begin{table*}
\centering
\caption{Statistics of benchmark datasets for table region detection.}
\begin{tabular}{|c|c|c|c|c|}
  \hline
 {\bf Dataset} & {\bf \# Training files}& {\bf \# Testing PDF files} & {\bf \# Training Lines} & {\bf \# Testing Lines} \\
 \hline
 \hline
  {\it ICDAR 2013}  & 50 & 17 & 804 & 224\\
  {\it ACL Anthology} & 9,280& 186 & 357,892 & 346\\
  \hline
\end{tabular}

\end{table*}

\subsection{Region detection}
Suppose that we have $n$ examples in the weakly labeled training dataset. Each example is a line of ``rich characters'' mapped into a feature vector ${\bf x}$ along with its weak label $y$. We further use $x_{NAM}$, $x_{PTD}$ and $x_{NEP}$ to denote the three features, i.e. {\it Normalized Average Margin (NAM)}, {\it POS Tag Distribution (PTD)} and {\it Named Entity Percentage (NEP)}, respectively. Hence, each training example can be represented as $({\bf x}^{(i)}, y^{(i)})$, in which ${\bf x}^{(i)} = \langle x^{(i)}_{NAM}, x^{(i)}_{PTD}, x^{(i)}_{NEP} \rangle$ and $(i)$ shows the index.

Here we use three canonical classifiers, i.e. {\it Logistic Regression}, {\it Support Vector Machine} and {\it Naive Bayes}, fed by the training examples above to decide whether a line of ``rich characters'' provided by {\it Apache PDFBox} belongs to a table or not, and explain the details about how we model the classifiers based on the features and the weak labels from the corpora we have constructed:

\begin{itemize}
\item {\it Logistic Regression\footnote{To implement the classifier, we integrate the
LIBLINEAR: \url{http://liblinear.bwaldvogel.de/} into our system.}} ({\it LR}) assumes that we can score the $i$-th example to indicate whether it belongs to a table or not, by approximate its score as a linear function of the feature vector ${\bf x}^{(i)}$:
\begin{equation}
\begin{split}
& ~~~~ s_{{\bf \theta}}({\bf x}^{(i)})\\
& = {\bf \theta}^{T} {\bf x}^{(i)} \\
& = \theta_{NAM} x^{(i)}_{NAM} + \theta_{PTD} x^{(i)}_{PTD} + \theta_{NEP} x^{(i)}_{NEP} + \theta_0,
\end{split}
\end{equation}

where the ${\bf
\theta}$ represents the vector of parameters along with the features.
Then the classifier chooses the {\it sigmoid function} which maps the score into $[0.0, 1.0]$, to show the probability of the feature vector ${\bf x}^{(i)}$ extracted from a table:
\begin{equation}
Pr(y^{(i)} = 1| {\bf x}^{(i)}) = \frac{1}{1 + e^{-s_{\theta}({\bf x}^{(i)})}},
\end{equation}
otherwise
\begin{equation}
Pr(y^{(i)} = 0| {\bf x}^{(i)}) = \frac{1}{1 + e^{s_{\theta}({\bf x}^{(i)})}}.
\end{equation}

The objective is to estimate the best parameter vector ${\bf
\theta}$ via maximizing the log-likelihood of all training examples:
\begin{equation}
\argmax_{\theta} ~~~ \log \prod_{i = 1}^{n} {Pr(y^{(i)}|{\bf x} ^ {(i)})}
\end{equation}

\item {\it Support Vector Machine\footnote{LIBSVM: \url{http://www.csie.ntu.edu.tw/~cjlin/libsvm/} is the well-known open-source software that can be leveraged by {\it PdfExtra}.} (SVM) }  enhances the hypothesis of linear combination which is illustrated by Equation (1), by defining the functional margin $\gamma$ given a training example $({\bf x}^{(i)}, y^{(i)})$:
\begin{equation}
 \gamma^{(i)} = y^{(i)}( {\bf w} ^ T{\bf x}^{(i)} + b),
\end{equation}
where $y^{(i)} = \{+1, -1\}$, ${\bf w}$ is the vector of weights, and $b$ is the bias.
Among all of them, we use $\gamma'$ to denote the minimum margin:
\begin{equation}
\gamma' = \min_{i = 1, ..., n}\gamma^{(i)}.
\end{equation}
The objective shown as follows:
\begin{equation}
\begin{split}
& \max_{\gamma, {\bf w}, b} ~~~\frac{\gamma}{||{\bf w}||}\\
& s.t. ~~~~y^{(i)}( {\bf w} ^ T{\bf x}^{(i)} + b) \geq \gamma, i = 1, ..., n\\
\end{split}
\end{equation}
results in a classifier that separates the positive and the negative training examples with a ``gap''.

\item {\it Naive Bayes\footnote{We adopt \url{https://github.com/ptnplanet/Java-Naive-Bayes-Classifier} to implement the Naive Bayes classifier.} (NB)} is different from the two classifiers mentioned above. Firstly, it requires discrete variables as features, and we need to map $x_{NAM}$ and $x_{NEP}$ which are originally described by continuous variables, into discrete variables\footnote{$x_{NAM}$ generally ranges from $0.0$ to $1.0$. We set a step size that equals to 0.2 to map the continuous variables. For example, if $ 0.0 \leq x_{NAM} < 0.2$, then $x_{NAM} = 1$, and so on.}. Secondly, rather than directly modeling $Pr(y|{\bf x})$ as two discriminative models mentioned above, {\it Naive Bayes} is a classical generative model which attempts to describe the joint probability of ${\bf x}$ and $y$, i.e. $Pr({\bf x}, y)$:

\begin{equation}
    Pr({\bf x}, y) = Pr({\bf x}|y)Pr(y).
\end{equation}

We derive $Pr(y|{\bf x})$ from $Pr(y, {\bf x})$ based on the {\it Bayes Rule}, and choose the value of $y$ with higher probability of $Pr(y|{\bf x})$ as the result of prediction. Given a testing example ${\bf x}^{(j)}$, we use the subsequent equation to predict the result:
\begin{equation}
\begin{split}
  & ~~ ~~\argmax_{y} Pr(y|{\bf x}^{(j)})\\
  & = \argmax_{y} \frac{Pr({\bf x}^{(j)}|y)Pr(y)}{Pr({\bf x}^{(j)})}\\
  & = \argmax_{y} Pr({\bf x}^{(j)}|y)Pr(y).\\
\end{split}
\end{equation}
The key assumption of {\it Naive Bayes} is that all the features are independent from each other given $y$:
\begin{equation}
Pr({\bf x}^{(j)}| y) = Pr(x_{NAM}^{(j)}| y )Pr( x_{PTD}^{(j)} | y)Pr(x_{NEP}^{(j)}|y).
\end{equation}
Therefore, we believe it will behave differently from the other classifiers.
\end{itemize}

\section{Experiment}
We set experiments that conduct comparison between our paradigm and the state-of-the-art {\it Heuristics} approach \cite{klampfl2014comparison} evaluated on two datasets, i.e. {\it ACL Anthology} and {\it ICDAR 2013 Table Competition}, with standard metrics.
\subsection{Datasets}
We prepare two datasets from different domains. The dataset\footnote{\url{http://www.tamirhassan.com/files/icdar2013-competition-dataset-with-gt.zip}} of {\it ICDAR 2013 Table Competition} is the benchmark in which there are 67 ground-truth PDF documents of U.S. and E.U. governments. The size of the other {\it ACL Anthology} dataset is much larger, which contains 9,466 academic articles from the year 2000 to 2015. It covers more than 10 top-tier conferences related to {\it Computational Linguistics}, such as {\it ACL, EMNLP, COLING, NAACL}, etc.
Table 1 shows the statistics of {\it ICDAR 2013} and {\it ACL Anthology} datasets for evaluation.
\begin{itemize}
  \item {\it ICDAR 2013}: We divide the dataset into two parts. 75\% files (50 documents) are used as training examples, and 25\% files (17 documents) are prepared for testing. After processed by the {\it Heuristic annotation}, we get 804 lines left for training. And we manually annotate 224 lines from 17 testing documents for testing.
  \item {\it ACL Anthology}: The paper published in 2015 are kept, and we label 346 lines of them as the ground-truth examples for testing. In addition, we gain 357,892 lines from 9,280 academic articles as the weakly labeled training examples.
\end{itemize}
\subsection{Metrics}
Since we regard the table region detection as binary classification problems, several standard metrics, such as {\it Accuracy}, {\it Precision}, {\it Recall}, {\it F1-measure}, are adopted for evaluating the performances. Each ground-truth line for testing is classified based on its features, and the output labels will be $+1$ or $-1$. As shown in Figure 2, anyone of the testing examples will fall into one of the four cells, i.e. {\it True Positive}. For instance, if a system assigns the positive label ($+1$) to a ground-truth testing line which should be regarded as a negative example, that is a false positive ($FP$).
\begin{figure}[!htp]
\centering
\includegraphics[width=0.5\textwidth]{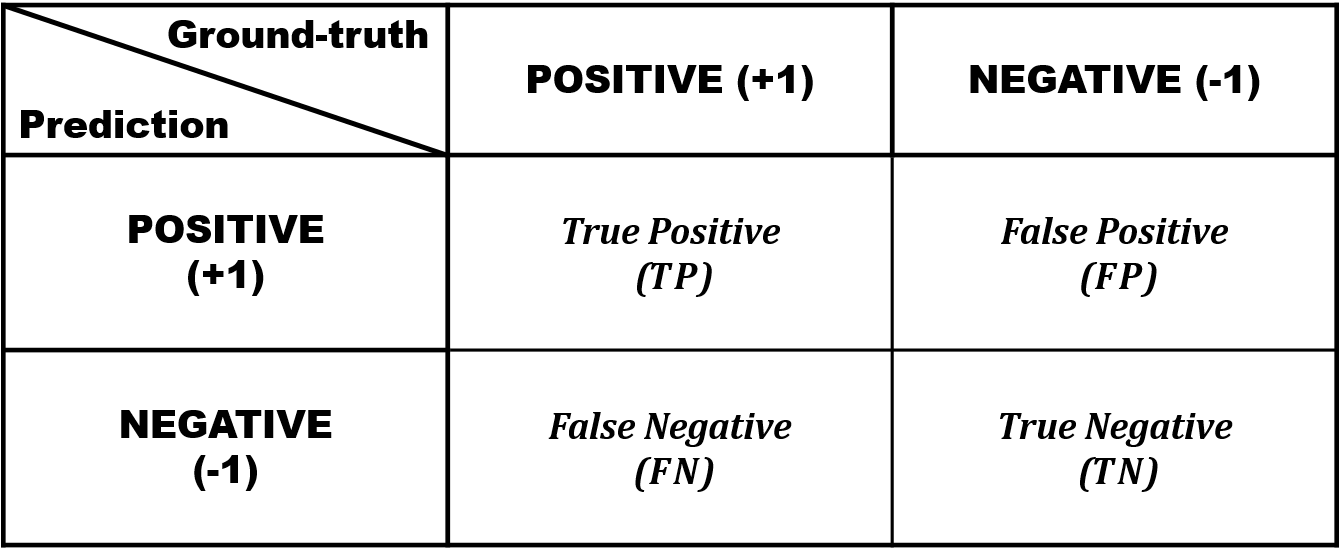}\\
\caption{$2 \times 2$  evaluation matrix for binary classification. }
\end{figure}

\begin{itemize}
  \item {\it Accuracy} is a metric to measure the overall performance of binary classification. It concerns about all the testing examples, including the positives and the negatives, and indicates the proportion of lines that are made correct predictions. Therefore,
\begin{equation}
Accuracy = \frac{\#(TP) + \#(TN)}{\#(TP) + \#(FP) + \#(FN) + \#(TN)}.
\end{equation}
  \item {\it Precision} and {\it Recall} are a pair of metrics that focus on the positive ground-truth lines. Specifically, {\it precision} represents the proportion of correct examples regarded as the positives, i.e.,
  \begin{equation}
Precision = \frac{\#(TP)}{\#(TP) + \#(FP)},
\end{equation}
and {\it recall} concerns about the proportion of positive predictions within all positive ground-truth examples:
\begin{equation}
Recall = \frac{\#(TP)}{\#(TP) + \#(FN)}.
\end{equation}

  \item {\it F1-measure} is a trade-off between {\it precision} and {\it recall}, which measures the harmonic mean of the two metrics above:
  \begin{equation}
F1{\textit -}measure = \frac{2 \times Precision \times Recall}{Precision + Recall}.
\end{equation}

\end{itemize}
\subsection{Performances}
We use the four metrics above to measure the performances of our system {\it PdfExtra}, compared with the state-of-the-art approach {\it Heuristics} \cite{klampfl2014comparison}. Both of them are evaluated by two benchmark datasets, i.e. {\it ICDAR 2013} and {\it ACL Anthology}. Table 2 and 3 show the results of the experiments, and we find out that {\it PdfExtra} achieves significant improvements beyond the latest approach.

\begin{table*}
\centering
\caption{Performance comparison on {\it ICDAR 2013} testing set, fed by {\it ICDAR 2013} training examples.}

\begin{tabular}{|c|c|c|c|c|}
  \hline
 {\bf Approach} & {\bf Accuracy}& {\bf Precision} & {\bf Recall} & {\bf F1-measure} \\
 \hline
 \hline
  {\it Heuristics} & 0.5491 & 0.5946 & 0.3826 & 0.4656\\
  {\it PdfExtra} & {\bf 0.6607} & {\bf 0.7407} & {\bf 0.5217} & {\bf 0.6122}\\
  \hline
\end{tabular}
\end{table*}

\begin{table*}
\centering
\caption{Performance comparison on {\it ACL Anthology} testing set, fed by {\it ACL Anthology} training examples.}

\begin{tabular}{|c|c|c|c|c|}
  \hline
 {\bf Approach} & {\bf Accuracy}& {\bf Precision} & {\bf Recall} & {\bf F1-measure} \\
 \hline
 \hline
  {\it Heuristics}  & 0.5665 & 0.5660 & 0.3659 & 0.4444\\
  {\it PdfExtra} & {\bf 0.7948} & {\bf 0.7385} & {\bf 0.8780} & {\bf 0.8022}\\
  \hline
\end{tabular}

\end{table*}

\begin{table*}
\centering
\caption{Performance comparison with different combinations of features on {\it ICDAR 2013} testing set, fed by {\it ICDAR 2013} training examples.}
\begin{tabular}{|c|c|c|c|c|}
  \hline
 {\bf Approach} & {\bf Accuracy}& {\bf Precision} & {\bf Recall} & {\bf F1-measure} \\
 \hline
 \hline
  {\it Heuristics}  & 0.5491 & 0.5946 & 0.3826 & 0.4656\\
   {\it PdfExtra (NAM)}  & 0.5134 &  0.5134 & {\bf 1.0000} & 0.6785\\

  {\it PdfExtra (NAM + PTD)}  & {\bf 0.7321} & {\bf 0.7835} & 0.6609 & {\bf 0.7170}\\

  {\it PdfExtra (NAM + PTD + NEP)}   & 0.6607 & 0.7407 & 0.5217 & 0.6122\\
  \hline
\end{tabular}
\end{table*}

\begin{table*}
\centering
\caption{Performance comparison with different combinations of features on {\it ACL Anthology} testing set, fed by {\it ACL Anthology} training examples.}
\begin{tabular}{|c|c|c|c|c|}
  \hline
 {\bf Approach} & {\bf Accuracy}& {\bf Precision} & {\bf Recall} & {\bf F1-measure} \\
 \hline
 \hline
  {\it Heuristics}  & 0.5665 & 0.5660 & 0.3659 & 0.4444\\
   {\it PdfExtra (NAM)}  & 0.4740 & 0.4740 & {\bf 1.0000} & 0.6431\\

  {\it PdfExtra (NAM + PTD)}  & 0.7312 & 0.6564 & 0.9085 & 0.7621\\

  {\it PdfExtra (NAM + PTD + NEP)}  & {\bf 0.7948} & {\bf 0.7385} & 0.8780 & {\bf 0.8022}\\
  \hline
\end{tabular}
\end{table*}

\begin{table*}
\centering
\caption{Performance comparison with different classifiers on {\it ICDAR 2013} testing set, fed by {\it ICDAR 2013} training examples.}
\begin{tabular}{|c|c|c|c|c|}
  \hline
 {\bf Approach} & {\bf Accuracy}& {\bf Precision} & {\bf Recall} & {\bf F1-measure} \\
 \hline
 \hline
  {\it Heuristics}  & 0.5491 & 0.5946 & 0.3826 & 0.4656\\
  {\it PdfExtra (NB)}  & {\bf 0.7902} & {\bf 0.7464} & {\bf 0.8957} & {\bf 0.8142}\\

  {\it PdfExtra (LR)}  & 0.6071 & 0.6956 & 0.4174 & 0.5217\\

  {\it PdfExtra (SVM)} & 0.6607 & 0.7407 & 0.5217 & 0.6122\\

  {\it PdfExtra} & 0.6607 & 0.7407 & 0.5217 & 0.6122\\
  \hline
\end{tabular}
\end{table*}

\begin{table*}
\centering
\caption{Performance comparison with different classifiers on {\it ACL Anthology} testing set, fed by {\it ACL Anthology} training examples.}
\begin{tabular}{|c|c|c|c|c|}
  \hline
 {\bf Approach} & {\bf Accuracy}& {\bf Precision} & {\bf Recall} & {\bf F1-measure} \\
 \hline
 \hline
  {\it Heuristics}  & 0.5665 & 0.5660 & 0.3659 & 0.4444\\
  {\it PdfExtra (NB)}  & 0.7861 & 0.7206 & {\bf 0.8963} & 0.7989\\

  {\it PdfExtra (LR)} & 0.7948 &  0.7385 & 0.8780 & 0.8022\\

  {\it PdfExtra (SVM)}  & 0.7919 & 0.7347 & 0.8780 & 0.8000\\

  {\it PdfExtra} & {\bf 0.7948} & {\bf 0.7385} & 0.8780 & {\bf 0.8022}\\

  \hline
\end{tabular}

\end{table*}

\begin{table*}
\centering
\caption{Cross-domain performance comparison on {\it ICDAR 2013} testing set, fed by {\it ACL Anthology} training examples.}
\begin{tabular}{|c|c|c|c|c|}
  \hline
 {\bf Approach} & {\bf Accuracy}& {\bf Precision} & {\bf Recall} & {\bf F1-measure} \\
 \hline
 \hline
  {\it Heuristics}  & 0.5491 & 0.5946 & 0.3826 & 0.4656\\
  {\it PdfExtra (ICDAR)} & 0.6607 & 0.7407 & 0.5217 & 0.6122\\

  {\it PdfExtra (ACL)}  & {\bf 0.7803} & {\bf 0.7683} & {\bf 0.7683} & {\bf 0.7683}\\
  \hline
\end{tabular}

\end{table*}

\section{Discussion}
To deeply analyze the paradigm we have proposed, we discuss the factors that may impact the performance from three perspectives:
\subsection{Impact of features}
We try different combinations of features. They are the layout feature only ({\it NAM}), the layout with part-of-speech feature ({\it NAM} + {\it PTD}) and the combination of all the features ({\it NAM} + {\it PTD} + {\it NEP}).
We keep collaboratively using the three classification models to vote the final prediction. Both of Table 4 and 5 demonstrate that pure layout feature does not perform well on detecting the table region, as shown by the experimental results of state-of-the-art {\it Heuristics}\cite{klampfl2014comparison} and {\it PdfExtra (NAM)}. For {\it ICDAR 2013} dataset, the best feature combination is {\it NAM} + {\it PTD}. And the other empirical study displays that the feature combination of {\it NAM} + {\it PTD} + {\it NEP} leads to the best performance on {\it ACL Anthology} dataset.

\subsection{Impact of classifiers}

Besides the combinations of features, three classifiers may also perform variously, due to their different hypotheses of mathematical modeling.
Therefore, we map both {\it ICDAR 2013} and {\it ACL Anthology} datasets to the same feature combination ({\it NAM} + {\it PTD} + {\it NEP}) schema, and iteratively select an individual classifier, such as Naive Bayes ({\it PdfExtra(NB)}), Logistic Regression ({\it PdfExtra(LR)}) or Support Vector Machine ({\it PdfExtra(SVM)}), to compare with
the voting version ({\it PdfExtra}).
They are the layout feature only ({\it NAM}), the layout with part-of-speech feature ({\it NAM} + {\it PTD}) and the combination of all the features ({\it NAM} + {\it PTD} + {\it NEP}).
Table 6 and 7 show the performances on {\it ICDAR 2013} and {\it ACL Anthology} datasets respectively, and {\it Naive Bayes} classifier behaves stably on the two benchmark datasets.

\subsection{Impact of domains}
The most significant perspective of our new paradigm that needs to be discussed, is the evaluation on cross-domain datasets. It directly reflects the generality of a paradigm. If it could only outperform the state-of-the-art approaches when trained and tested by the PDF documents in the same or specific domain, the paradigm would still be a trial version that make minor contributions on the research of table region detection. Therefore, An experiment is set in which we feed the training examples of {\it ACL Anthology} dataset to our model, and test the performance on the testing set of {\it ICDAR 2013}. Fortunately, testing on files of {\it ICDAR 2013} achieves comparable performance with testing on {\it ACL Anthology} dataset. Moreover, {\it PdfExtra (ACL)} shows the better capability on detecting tables on government documents after trained by academic articles. The reason why our paradigm can handle cross-domain files, is that all the features and classifiers we leverage are independent from the layouts, and even the contents within diverse PDF documents.

\section{Conclusion}
In this paper, we have contributed a novel paradigm for detecting the region of tables within PDF documents. It absorbs superiorities from both supervised and unsupervised approaches, and firstly covers almost tens of thousands PDF documents in a different domain. To be specific, it leverages different supervised learning models to adapt varies layouts and linguistic features within tables from large-scale PDF files, but costs free on labeling training corpus. We integrate the paradigm into our system {\it PdfExtra} which enhances the off-the-shelf software {\it Apache PDFBox} to predict whether a text line belongs to a table or not. Three classification models have been evaluated, which are {\it Logistic Regression}, {\it Support Vector Machine}, and {\it Naive Bayes}. We find out that {\it Naive Bayes} performs stable prediction on both two benchmark datasets, and linguistic features bring a great leap forward on the performance. What's more, we prove that our paradigm is robust to table detection on open-domain PDF documents.

However, the idea of weakly labeled paradigm can not avoid bringing noise into training data which impacts the performance of system. In the future, we look forward to exploring the correlation between tables within the same article to filter out the faults.


\bibliographystyle{IEEEtran}
\bibliography{ref2}



%
%
%

\end{document}